%% file: paper.tex
\documentclass[runningheads]{llncs}

\RequirePackage[hyphens]{url} 
\PassOptionsToPackage{hyphens}{url}\usepackage{hyperref} 

\usepackage[T1]{fontenc}
\usepackage{graphicx}
\usepackage{stackengine}
\usepackage{layouts}

\usepackage{array} 
\usepackage{pbox} 
\usepackage{amsmath} 
\usepackage{amssymb}
\usepackage[super]{nth}
\usepackage{marvosym} 
\usepackage{capt-of}
\usepackage{subcaption} 

\usepackage{booktabs}
\usepackage{tabularray}
\usepackage{multirow}
\usepackage{comment}

\usepackage[table]{xcolor}
\usepackage{pgf}
\usepackage{collcell}
\usepackage{layouts}
\usepackage{float}
\usepackage{adjustbox}

\newcommand*{\MinNumber}{40}%
\newcommand*{\MaxNumber}{200}%

\newcommand{\ApplyGradient}[1]{%
  \pgfmathsetmacro{\PercentColor}{100*(#1-\MinNumber)/(\MaxNumber-\MinNumber)}%
  \edef\x{\noexpand\cellcolor{black!\PercentColor}}\x\textcolor{black}{#1}%
}
\newcolumntype{R}{>{\collectcell\ApplyGradient}{c}<{\endcollectcell}}

\begin{document}


\title{A Real-Time Digital Twin for Type 1 Diabetes using Simulation-Based Inference}

\titlerunning{A Real-Time Digital Twin for T1D using Simulation-Based Inference}

\author{
    Trung-Dung Hoang \inst{1,2,3}\textsuperscript{(\Letter)} \and
    Alceu Bissoto\inst{1,2,3} \and
    Vihangkumar V. Naik\inst{1,2,3} \and 
    Tim~Flühmann\inst{1,2,3} \and
    Artemii Shlychkov\inst{1,2,3,4} \and
    Jose Garcia-Tirado\inst{1,2,3} \and
    Lisa~M.~Koch\inst{1,2,3}\textsuperscript{(\Letter)}
}
\authorrunning{Hoang et al.}
\institute{
    University of Bern, Switzerland \and
    Department of Diabetes, Endocrinology, Nutritional Medicine and Metabolism, Inselspital, Bern University Hospital, University of Bern, Switzerland \and
    Diabetes Center Berne, Switzerland \and
    University of Tübingen, Germany \\
    \email{\{trung.hoang,lisa.koch\}@unibe.ch}
}
\maketitle

\setcounter{footnote}{0}
\begin{abstract}

Accurately estimating parameters of physiological models is essential to achieving reliable digital twins. For Type 1 Diabetes, this is particularly challenging due to the complexity of glucose–insulin interactions.
Traditional methods based on Markov Chain Monte Carlo struggle with high-dimensional parameter spaces and fit parameters from scratch at inference time, making them slow and computationally expensive.
In this study, we propose a Simulation-Based Inference approach based on Neural Posterior Estimation to efficiently capture the complex relationships between meal intake, insulin, and glucose level, providing faster, amortized inference.
Our experiments demonstrate that SBI not only outperforms traditional methods in parameter estimation but also generalizes better to unseen conditions, offering real-time posterior inference with reliable uncertainty quantification.\footnote{Code available at \url{https://github.com/mlm-lab-research/SBI_T1D}}

\end{abstract}

\input{01-introduction}

\input{02-background}

\input{03-method}

\input{04-exp_main}

\input{05-discussion}

\subsubsection{\ackname} 

This project was supported by the Diabetes Center Berne and strategic funding of the medical faculty of the University of Bern. Calculations were performed on UBELIX, the HPC cluster at the University of Bern.

\bibliographystyle{splncs04_without_urls}
\bibliography{references.bib}

\end{document}

%% file: 01-introduction.tex
\section{Introduction}

Type 1 diabetes (T1D) affects more than 9 million people worldwide \cite{T1DIndex2025}. T1D is an autoimmune condition resulting in an absolute deficiency in insulin, a hormone essential for allowing glucose to enter cells and produce energy. Individuals with T1D therefore require frequent insulin injections and close monitoring of their glucose levels, e.g., with wearable continuous glucose monitoring (CGM) devices. Maintaining healthy glucose levels through carefully administered insulin is crucial: too much insulin can lead to life-threatening hypoglycaemia, while too little insulin results in increased blood glucose levels, which can cause many serious long-term micro- and macrovascular complications.

The dynamics between glucose, meal intake, and insulin over time can be described by complex physiological models consisting of systems of differential equations \cite{dalla2007meal,foster1973short,bergman1979quantitative}, where model parameters (e.g., insulin sensitivity) can be highly patient-specific. Identifying patient-specific parameters from observed data allows creating a digital twin (DT) of the individual’s metabolic system. This could improve treatment planning, prediction, and real-time adaptation, where the DT is continuously updated with patient data, allowing the system to better respond to daily variations in meals, insulin dosing, and insulin sensitivity \cite{Kovatchev2025}.

Despite the promise of DTs, parameter estimation in T1D models is a challenging inverse problem due to complex, nonlinear relationships among system components.
The current gold standards (e.g., Markov Chain Monte Carlo (MCMC))\cite{cappon2023replaybg,COLMEGNA2020104605,haidar2013stochastic,visentin2016one} attempt to directly fit the physiological model to individual data.
This means they require a separate optimization or sampling process for each new observation (non-amortized), making them computationally expensive, limited to low-dimensional problems, and unsuitable for real-time applications.
Furthermore, an often overlooked challenge is inferring the system's initial conditions. Many current approaches are limited to a steady-state initialisation. In the case of T1D, this means no meals or insulin could be administered for at least four hours before the observation window starts.
If the assumption is not met, the inferred parameters may overcompensate for the wrong initial conditions, resulting in inaccurate parameter estimates and inadequate use of the DT. 

To address these challenges, Simulation-Based Inference (SBI) offers a more flexible and scalable approach that uses neural networks for probabilistic inference \cite{Cranmer_2020}. SBI has been successfully applied to inverse problems in different fields, such as astrophysics \cite{PhysRevLett.127.241103,bhardwaj2024peregrinesequentialsimulationbasedinference,Mishra_Sharma_2022} and cardiovascular simulations \cite{simulation-based-inference}. 
SBI requires only data simulated from physiological models to train the neural network,
enabling it to work directly with any simulator.
Furthermore, by leveraging Normalizing Flows \cite{normalizingflow} in particular the Neural Posterior Estimation (NPE) framework \cite{NIPS2017_addfa9b7,NIPS2016_6aca9700}, SBI can efficiently model complex, high-dimensional posterior distributions, allowing inference over a larger set of parameters.

In this work, we present the first application of SBI to CGM data and T1D models, and compare it to the previous state-of-the-art for parameter estimation in T1D models \cite{cappon2023replaybg}. Our approach directly tackles both the computationally expensive nature of state-of-the-art techniques and the challenging inference of the initial conditions.
Our method is computationally cheap. Once trained, the neural network can rapidly infer parameters for new patients without running a separate optimization or sampling routine. 
We also show how SBI can be naturally extended to estimate initial conditions of physiological models, such that our proposed approach works reliably in more realistic scenarios beyond steady-state conditions.

%% file: 02-background.tex
\section{Background}
\subsection{Physiological T1D models}

\label{sec:t1d_model}

Physiological T1D models aim to mathematically represent how the body regulates blood glucose through the dynamics of insulin and glucose in people with T1D. In this work, we use a simplified version \cite{cappon2023replaybg} of the UVA/Padova simulator \cite{visentin2018uva}, a widely accepted and clinically validated physiological model of T1D. The model is divided into four main subsystems:

\textbf{Subcutaneous insulin absorption} models how delivered insulin $I$ reaches the plasma after injection. It first transitions to non-monomeric insulin $I_{sc1}$, then to monomeric insulin $I_{sc2}$, and finally to plasma insulin $I_p$:

\begin{equation*}
\begin{cases}
\dot{I}_{sc1}(t) = -k_d \cdot I_{sc1}(t) + I(t - \beta)/V_I \\
\dot{I}_{sc2}(t) = k_d \cdot I_{sc1}(t) - k_{a2} \cdot I_{sc2}(t) \\
\dot{I}_p(t) = k_{a2} \cdot I_{sc2}(t) - k_e \cdot I_p(t)
\end{cases}
\end{equation*}
Here, $k_d$ is the transition rate from non-monomeric to monomeric insulin, $k_{a2}$ is the absorption rate into plasma, $k_e$ is the clearance rate, $V_I$ is the insulin distribution volume, and $\beta$ is the insulin appearance delay.

\textbf{Oral glucose absorption} models how ingested carbohydrates $CHO$ transition through the digestive system. Glucose enters the stomach as solid state ($Q_{sto1}$), converts to liquid state ($Q_{sto2}$), moves to the intestine ($Q_{gut}$), and finally appears in plasma as $Ra$ (rate of glucose appearance):

\begin{equation*}
\begin{cases}
\dot{Q}_{sto1}(t) = -k_{empt} \cdot Q_{sto1}(t) + CHO(t) \\
\dot{Q}_{sto2}(t) = k_{empt} \cdot Q_{sto1}(t) - k_{empt} \cdot Q_{sto2}(t) \\
\dot{Q}_{gut}(t) = k_{empt} \cdot Q_{sto2}(t) - k_{abs} \cdot Q_{gut}(t) \\
Ra(t) = f \cdot k_{abs} \cdot Q_{gut}(t)
\end{cases}
\end{equation*}
Here, $k_{empt}$ is the gastric emptying rate, and $k_{abs}$ and $f$ are the intestinal absorption rate and fraction of glucose absorption through the intestinal wall, respectively.

\textbf{Glucose-insulin kinetics} models how glucose and insulin interact in the body. Plasma glucose $G$ is affected by insulin action $X$ and glucose appearance $Ra$. Insulin action $X$ depends on plasma insulin $I_p$, and interstitial glucose $IG$ reflects glucose transport between plasma and the interstitium:
\vspace{-0.05cm}
\begin{equation*}
\begin{cases}
\dot{G}(t) = -\rho(G) \cdot X(t) \cdot G(t) - SG \cdot [G(t) - G_b] + Ra(t)/V_G \\
\dot{X}(t) = -p_2 \cdot X(t) + p_2 \cdot SI \cdot (I_p(t) - I_{pb}) \\
\dot{IG}(t) = - (IG(t) - G(t))/\alpha
\end{cases}
\end{equation*}
Here, $SG$ is the fractional glucose effectiveness, quantifying how well glucose regulates itself by stimulating its uptake and suppress its production to maintain basal glucose levels $G_b$; $V_G$ is the glucose distribution volume; $SI$ is insulin sensitivity, measuring how effectively insulin lowers blood glucose; $I_{pb}$ is the basal insulin concentration; $p_2$ is the insulin action decay rate, and $\alpha$ is the plasma-to-interstitium transport delay.
The function $\rho(G)$ \cite{cappon2023replaybg} increases insulin action $X$ under hypoglycemia to better capture glucose dynamics in the low-glucose range.

\textbf{CGM sensor error} models the error introduced by the CGM sensor:

\begin{equation*}
CGM(t) = [(a_0 + a_1 \cdot t + a_2 \cdot t^2) \cdot IG(t) + b_0] + v(t)
\end{equation*}
where $a_0$, $a_1$, $a_2$, $b_0$ are sensor coefficients; $v(t) \sim N(0, \epsilon_v)$ is white noise.

Given meal inputs $CHO(t)$, insulin inputs $I(t)$, initial state values, and model parameters, the simulator generates CGM readings $CGM(t)$. Some parameters, such as $SI$, are highly patient-specific \cite{THOMSEN1997374}. Our goal is to estimate these individual-specific parameters from observed inputs and outputs.

\subsection{Simulation-Based Inference (SBI)}

Traditional Bayesian inference computes the posterior distribution of parameters $\boldsymbol{\theta}$ given observed data $\boldsymbol{y}$ as
$p(\boldsymbol{\theta} \mid \boldsymbol{y}) \propto p(\boldsymbol{y} \mid \boldsymbol{\theta}) \cdot p(\boldsymbol{\theta})$, where $p(\boldsymbol{y} \mid \boldsymbol{\theta})$ is the likelihood and $p({\boldsymbol{\theta}})$ is the prior.
In many scientific applications, including physiological T1D models, even though we can simulate $\boldsymbol{y}$ from the model given any $\boldsymbol{\theta}$, the likelihood function $p(\boldsymbol{y} \mid \boldsymbol{\theta})$ is intractable - we cannot evaluate it explicitly or compute its value efficiently. This happens when the simulator involves stochasticity, latent states, or complex numerical solvers. As a result, inference methods such as MCMC often assume a tractable form (e.g., a Gaussian likelihood), which may fail to capture the true complexity of the models.

Simulation-Based Inference (SBI) \cite{Cranmer_2020}, also known as likelihood-free inference, overcomes this by relying only on a simulator that can generate synthetic data $\boldsymbol{\tilde{y}} \sim p(\boldsymbol{y} \mid \boldsymbol{\theta})$ for any chosen parameter configuration $\boldsymbol{\theta}$. The goal is to learn the posterior distribution $p(\boldsymbol{\theta} \mid \boldsymbol{y})$ directly from these samples, without requiring an explicit likelihood function.
A powerful SBI approach is Neural Posterior Estimation (NPE)~\cite{NIPS2017_addfa9b7,NIPS2016_6aca9700}, which uses Conditional normalizing flows \cite{NIPS2016_6aca9700} to approximate the posterior. A normalizing flow $f_\phi$ \cite{normalizingflow} is an invertible neural network that transforms a simple distribution $p_Z$(e.g., a standard Gaussian) into a more complex one. In the conditional setting, the transformation depends on $\boldsymbol{y}$:
\[
\boldsymbol{\theta} = f_\phi(\boldsymbol{z}; \boldsymbol{y}), \quad \boldsymbol{z} \sim p_Z(\boldsymbol{z})
\]
The resulting density over $\boldsymbol{\theta}$ is given by the change-of-variables formula:
\[
q_\phi(\boldsymbol{\theta} \mid \boldsymbol{y}) = p_Z(\boldsymbol{z}) \left| \det \left( \frac{\partial f_\phi^{-1}(\boldsymbol{\theta}; \boldsymbol{y})}{\partial \boldsymbol{\theta}} \right) \right|
\] which is used to approximate $p(\boldsymbol{\theta} \mid \boldsymbol{y})$.
 
To train the flow, we maximize the conditional density $q_\phi(\boldsymbol{\theta^{(i)}} \mid \boldsymbol{\tilde{y}^{(i)}})$ for simulated pairs  $(\boldsymbol{\theta^{(i)}}, \boldsymbol{\tilde{y}^{(i)}})$. At inference, given observed $\boldsymbol{y}$, we can sample multiple $\boldsymbol{z} \sim p_Z(\boldsymbol{z})$ and apply the learned $f_\phi$ to obtain $\boldsymbol{\theta} = f_\phi(\boldsymbol{z}; \boldsymbol{y})$ from the approximate posterior.
This approach enables efficient parameter estimation in complex physiological models without any assumption about the likelihood function.

%% file: 03-method.tex
\section{SBI for T1D model identification}

\subsection{Parameters and prior distribution}
We apply SBI to estimate parameters of the T1D model described in Section~\ref{sec:t1d_model}. We focus on estimating a subset of eight physiological parameters, while keeping the remaining parameters fixed at population-level constants, following the state-of-the-art ReplayBG baseline, which uses MCMC as its backend \cite{cappon2023replaybg}:
\[
\boldsymbol{\theta} = [G_b, SG, p_2, k_{a2}, k_d, k_{empt}, SI, k_{abs}]
\]

We propose to treat the initial conditions as model parameters, enabling joint inference with physiological parameters via SBI. This helps avoid compensatory effects and bias caused by misspecified initial states. We consider the following nine initial state variables:
\[
\mathbf{x}_0 = [G(0), I_{sc1}(0), I_{sc2}(0), I_{p}(0), Q_{sto1}(0), Q_{sto2}(0), Q_{gut}(0), X(0), IG(0)]
\]
The full vector to be inferred is thus $\hat{\boldsymbol{\theta}} = [\boldsymbol{\theta}, \mathbf{x}_0] \in \mathbb{R}^{17}$. The observation vector $\boldsymbol{y} \in \mathbb{R}^{264}$ consists of CGM readings recorded every 5 minutes over 22 hours, as in the ReplayBG implementation \cite{cappon2023replaybg}.
The prior over $\hat{\boldsymbol{\theta}}$ is constructed as $p(\hat{\boldsymbol{\theta}}) = p(\boldsymbol{\theta}, \mathbf{x}_0) = p(\boldsymbol{\theta}) \cdot p(\mathbf{x}_0 \mid \boldsymbol{\theta})$. The marginal prior $p(\boldsymbol{\theta})$ follows the ReplayBG configuration. 
Since $p(\mathbf{x}_0 \mid \boldsymbol{\theta})$ has no closed-form, we propose sampling $\boldsymbol{x}_0$ by simulating 44 hours from a known steady state using the given $\boldsymbol{\theta}$ and extracting a random 22-hour window as $\boldsymbol{y}$. The initial nine state values from this shifted observation window are used as $\mathbf{x}_0$.
\vspace{-0.1cm}

\subsection{Training details}

To generate training data, we draw parameter vectors $\boldsymbol{\hat{\theta}}$ and, for each sample, simulate CGM observations using the T1D model with a fixed meal and insulin profile. We use rejection sampling and retain only the simulations with CGM outputs in the range $[40, 400]$ mg/dL until 5,000 valid samples are collected.

For the posterior estimator $f_\phi$, we use the implementation from the sbi library \cite{tejero-cantero2020sbi}. Specifically, we employ a Masked Autoregressive Flow (MAF) \cite{maf} composed of Masked Affine Autoregressive Transform layers, each conditioned on the observation $\boldsymbol{y}$. 
We train $f_\phi$ with the following configuration: batch size of 200, learning rate of $5\times10^{-4}$, gradient norm clipped at 5.0, and validation fraction of 10\%. Early stopping is triggered after 20 epochs of no improvement.

\subsection{Evaluation and baselines}

We evaluate our T1D digital twin on 50 simulated CGM trajectories, each generated from a different parameter vector drawn from the prior distribution. For each CGM trajectory, we use SBI to infer 1,000 posterior samples. We then assess the quality of these posterior estimates by comparing them against the ground truth parameters using four metrics.

The \textbf{absolute error} and \textbf{relative error} measure the absolute and relative difference between the median of the 1,000 posterior samples and the true value, respectively, while the \textbf{mean absolute deviation (MAD)} captures the average absolute difference across all posterior samples.
The \textbf{coverage} metric indicates whether the true values lie within the 2.5th and 97.5th percentiles of the posterior samples.
We also evaluate our approach's ability to reconstruct the observed CGM trajectory using \textbf{Mean Absolute Relative Difference (MARD)} and \textbf{Root Mean Square Error (RMSE)} between the observed and the reconstructed signal, averaged over the 50 test samples.

We compare our approach to two baselines: ReplayBG \cite{cappon2023replaybg} and a Maximum a posteriori (MAP) approach using parameter search algorithms to find a point estimate which maximizes the posterior $p(\boldsymbol{\theta}|\boldsymbol{y})$. MCMC in ReplayBG is configured with 10,000 burn-in and 5,000 main steps. For the MAP approach, we use the implementation provided in \cite{cappon2023replaybg}.

%% file: 04-exp_main.tex
\section{Results}
\subsection{SBI improves parameter estimation in T1D digital twins}

\begin{figure}[H]
    \centering
    \includegraphics[width=\textwidth]{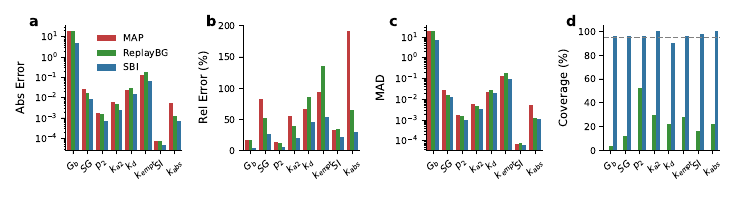}
    \caption{SBI outperforms baseline methods: \textbf{a-c)} Parameter estimation error and \textbf{d)} coverage of inferred parameter posteriors (dashed line shows 95\% coverage).}
    \label{fig:param_eval}
\end{figure}

SBI consistently led to the lowest parameter estimation errors (Fig.~\ref{fig:param_eval}\,\textbf{a,b,c}).
For basal glucose ($G_b$), a crucial parameter for T1D management, SBI attained an error of $5.04\,\mathrm{mg/dl}$ (4.32\% relative error) and a MAD of $7.19\,\mathrm{mg/dl}$, compared to ReplayBG's $19.12\,\mathrm{mg/dl}$ (16.24\%) and $19.15\,\mathrm{mg/dl}$, respectively.
The smaller errors of SBI could be attributed to its ability to learn complex distributions and estimate the initial conditions, whereas ReplayBG fixed the initial conditions at steady state and had to compensate the parameters to fit the observation.
By jointly estimating initial conditions and parameters, SBI could more accurately capture the true system dynamics.

\begin{figure}[t]
    \centering
    \includegraphics[width=0.73\linewidth]{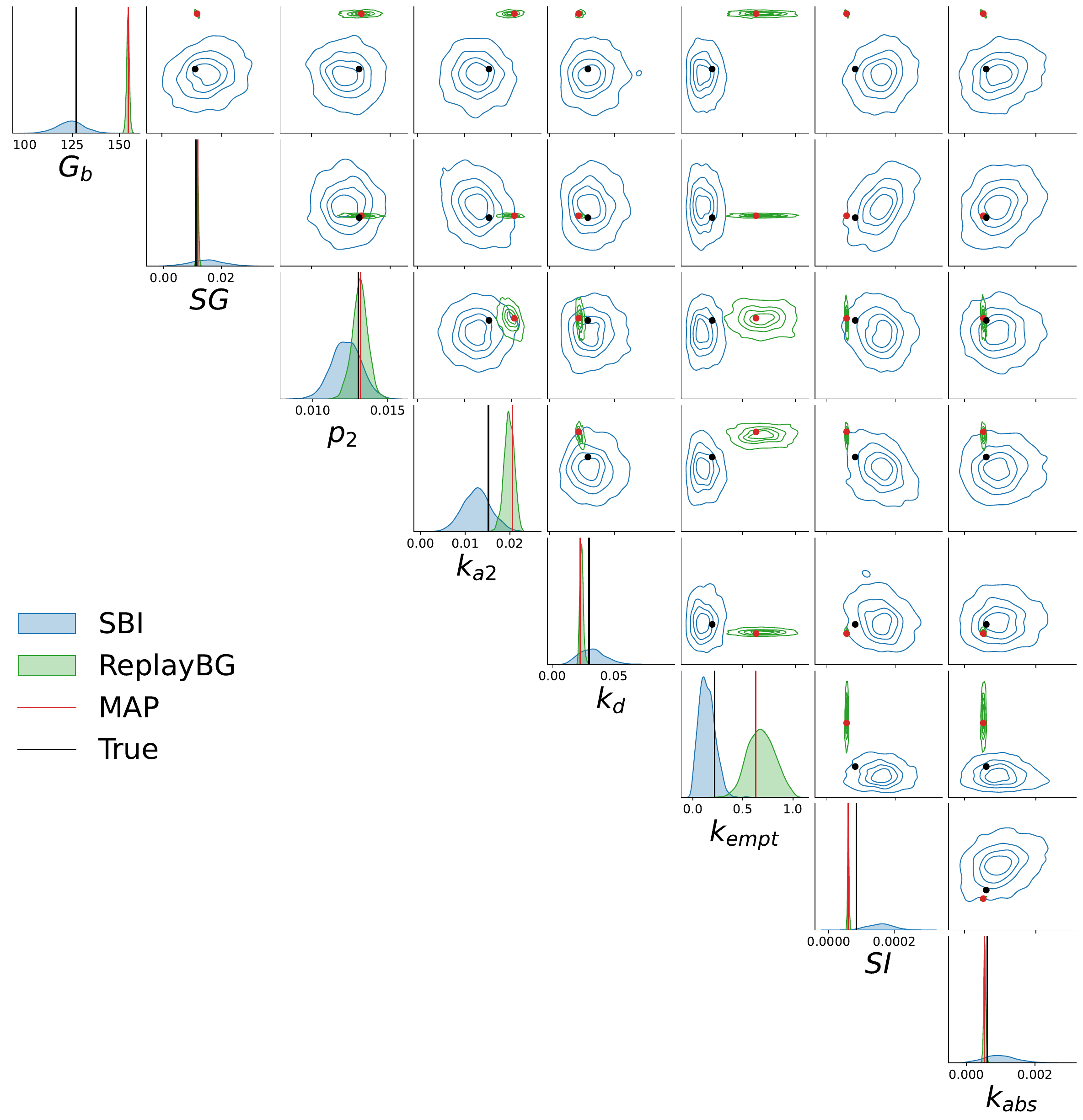}
    \caption{Visualization of posterior distributions of a single test case. SBI posteriors (blue) cover the true value (black) well, while MCMC posteriors (green) tend to be narrower and overconfident around the MAP estimate (red).}
    \label{fig:posterior_viz}
\end{figure}

The uncertainty quantification obtained through SBI's full posterior estimates led to large coverage for all parameters (Fig.~\ref{fig:param_eval}\,\textbf{d}), with an average coverage of 96.5\% within the 95\% credible interval, indicating well-calibrated uncertainty. In contrast, the MCMC posteriors were consistently overconfident, which can cover the true values in only 23.25\% of cases on average.
These results are further supported qualitatively in Fig.~\ref{fig:posterior_viz}, where posteriors of SBI covered the true parameters best. The MAP baseline provided only a point estimate without uncertainty quantification, making coverage calculation inapplicable.
\vspace{-0.2cm}
\vspace{-0.5cm}

\subsection{What-if scenario simulation: SBI generalises to unseen scenarios}

A key motivation for using DTs is to explore counterfactual or hypothetical scenarios that cannot be directly observed in real-world data. To evaluate generalization ability, we considered three settings where CGM signals were reconstructed from inferred parameters: (1) \textbf{In-sample reconstruction} - simulation using the same meal profile as in the observation; (2) \textbf{Out-of-sample: next day} - simulation extended to the next day; (3) \textbf{Out-of-sample: altered meals} - simulation with the meal profile modified. For SBI and ReplayBG, we used the median CGM signal simulated from 1,000 posterior samples as a replay signal, and for MAP, we used a single CGM signal simulated from the point estimate.

\begin{figure}[H]
    \centering
    \includegraphics[]{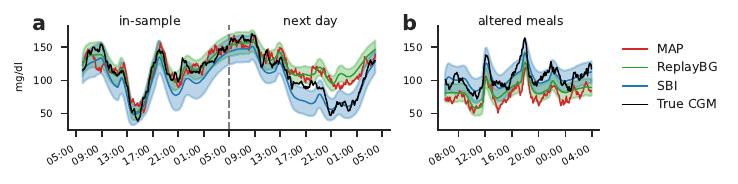}
    \caption{CGM reconstructions (colors depict different methods) for an example CGM signal (black). \textbf{a)} All methods matched the observed data on the first day (Setting 1), but only SBI accurately predicted the next day (Setting 2). \textbf{b)} SBI reconstructed the true CGM better than the baselines (Setting 3).}
    \label{fig:cgm_plot}
\end{figure}

ReplayBG outperformed SBI in the in-sample setting, as they are designed to fit the observed data closely (left columns in Tab.~\ref{tab:replay}). However, in out-of-sample scenarios, SBI consistently outperformed both baselines, achieving the lowest MARD and RMSE (middle and right columns in Tab.~\ref{tab:replay}). Fig.\,\ref{fig:cgm_plot} shows an example test case. While MAP and ReplayBG accurately reproduced the CGM on the first day, their replay signals deviated considerably on the second day. In contrast, SBI continued to follow the ground-truth trajectory thanks to SBI’s more accurate estimation of the physiological parameters, which is crucial for generalizing to unseen conditions.

\begin{table}[]\centering
\caption{Average MARD (\%) and RMSE ($\mathrm{mg/dL}$)
 across three settings.}\label{tab:replay}
\resizebox{\textwidth}{!}{

\begin{tabular}{ccccccc}\toprule
\multirow{2}{*}{} &\multicolumn{2}{c}{in-sample reconstruction} &\multicolumn{2}{c}{out-of-sample: next day} &\multicolumn{2}{c}{out-of-sample: altered meals} \\\cmidrule{2-7}
&MARD &RMSE &MARD &RMSE &MARD &RMSE \\\midrule

MAP &14.36 $\pm$ 6.00 &17.30 $\pm$ 5.43 &22.29 $\pm$ 15.47 &23.81 $\pm$ 12.12 &18.11 $\pm$ 7.84 &21.97 $\pm$ 8.45 \\
ReplayBG &\textbf{6.73 $\pm$ 2.82} &\textbf{8.31 $\pm$ 2.20} &20.20 $\pm$ 16.69 & 20.76 $\pm$ 10.83 & 15.05 $\pm$ 6.53 &17.83 $\pm$ 7.40 \\
\rowcolor[HTML]{EAEAEA}SBI &12.60 $\pm$ 6.19 &15.12 $\pm$ 6.79 &\textbf{14.84 $\pm$ 8.84} &\textbf{16.09 $\pm$ 7.54} &\textbf{12.68 $\pm$ 6.42} &\textbf{15.56 $\pm$ 8.60} \\

\bottomrule
\end{tabular}
}
\end{table}

\vspace{-1cm}

\subsection{SBI is suitable for real-time inference}

\vspace{-1cm}
\begin{table}[H]\centering
\caption{Inference and training time comparison across methods.}\label{tab:time}
\resizebox{0.5\textwidth}{!}{
\begin{tabular}{cccc}\toprule
Method &Inference Time (s) &Training Time (s) \\\midrule
MAP &51.46 $\pm$ 13.98 &0  \\
ReplayBG &2767.04 $\pm$ 258.36 &0  \\
\rowcolor[HTML]{EAEAEA}SBI &3.36 $\pm$ 0.52 &161.2 \\
\bottomrule
\end{tabular}
}
\end{table}
\vspace{-0.5cm}
Table~\ref{tab:time} highlights the efficiency of SBI in terms of inference speed and overall computational cost. Inference using SBI took 3.36 seconds on average, outperforming both baselines by a large margin. 
While MAP inference completed in less than a minute, it often yielded large errors. It also lacks uncertainty quantification, which is crucial for facilitating trust in the inference results.
ReplayBG can produce posterior samples but required approximately 45 minutes per sample. Although SBI needed 161.2 seconds to train, this one-time cost enables reuse without retraining, providing amortized inference. This makes SBI a great choice for tasks that need frequent or real-time inference, such as DTs, where they must quickly adapt to changes in the real system.
\vspace{-0.1cm}

%% file: 05-discussion.tex
\section{Discussion}

In this work, we propose using SBI for T1D digital twins. SBI is an efficient approach that enables accurate, efficient, and amortized parameter inference. Our extension of SBI to infer initial conditions enables it to operate beyond steady-state conditions. This makes it particularly well-suited for complex physiological systems where traditional methods often rely on those restrictive assumptions.

Our experiments demonstrate that SBI delivers accurate posterior estimates and significantly improves the quality of simulated CGM trajectories. Its fast, amortized inference opens the door for real-time DTs, which remain a challenge for existing methods. To further enhance the robustness of SBI, for future work, we plan to simulate a broader range of scenarios, such as varying meal and insulin profiles, to enrich the training data. Beyond glucose modeling, SBI offers opportunities for DTs in other domains, such as cardiac dynamics, making it a truly transformative tool in computational physiology.